\documentclass[letter,10pt,conference]{ieeeconf}
\IEEEoverridecommandlockouts
\usepackage{amsmath,amssymb,theorem}
\usepackage{graphicx}
\graphicspath{ {./images/} }
\usepackage{algorithm}
\usepackage{algpseudocode}
\usepackage{psfrag}
\usepackage{cite}
\usepackage{rotating}

\usepackage{color,xspace}
\usepackage{subfigure}
\usepackage{microtype}
\newtheorem{theorem}{Theorem}[section]
\newtheorem{lemma}[theorem]{Lemma}

\newtheorem{example}[theorem]{Example}

\newtheorem{proposition}[theorem]{Proposition}
\newtheorem{problem}[theorem]{Problem}

\newcommand{\real}{{\mathbb{R}}}

\newcommand{\GG}{{\mathcal{G}}}
\newcommand{\PP}{{\mathcal{P}}}

\newcommand{\VV}{{\mathcal{V}}}

\newcommand{\edges}{E}

\renewcommand{\epsilon}{\varepsilon}

\newcommand{\oprocendsymbol}{\hbox{$\bullet$}}
\newcommand{\oprocend}{\relax\ifmmode\else\unskip\hfill\fi\oprocendsymbol}

\newcommand{\longthmtitle}[1]{\mbox{}\textup{\textbf{(#1)}}}

\renewcommand{\theenumi}{(\arabic{enumi})}

\newcommand{\FOV}{\operatorname{FOV}}

\usepackage{xcolor}

\begin{document}

% \title{Chemistry, Fluid Mechanics, and Swarm Robotics: An Interdisciplinary Exploration}
%\title{Rethinking Swarm Robotics by Applying Concepts from Chemistry and Fluid Mechanics}
%\title{Swarm Mechanics: Treating Robot Swarms like a Fluid?}
%\title{\LARGE \bf Swarm Mechanics and Swarm Chemistry: \\ A Transdisciplinary Approach for Robot Swarms}

\title{\LARGE \bf Indirect Swarm Control: Characterization and Analysis of Emergent Swarm Behaviors}

%Controlling emergent behaviors through self-organizatino asfkjbWEGOIQEGFqewligweqglknvl,knsdvklnk,n

%\title{Converging Realms: Exploring the Interplay of Chemistry, Fluid Mechanics, and Swarm Robotics}

% Use-Inspired Emergent Behaviors in Robot Swarms

%
\author{Ricardo Vega\qquad Connor Mattson \qquad Daniel S. Brown \qquad Cameron Nowzari \thanks{R. Vega and C. Nowzari are
    with the Department of Electrical and Computer Engineering,
    George Mason University, Fairfax, VA 22030, USA, {\tt\small
      \{rvega7, cnowzari\}@gmu.edu}}
      \thanks{
      C. Mattson and D. S. Brown are
    with the Kahlert School of Computing,
    University of Utah, Salt Lake City, UT 84112,  USA, {\tt\small \{c.mattson,daniel.s.brown\}@utah.edu}}}
      
%      \author{Ricardo Vega\thanks{The authors are
%    with the Department of Electrical and Computer Engineering and Department of Computer Science,
%    George Mason University, Fairfax, VA 22030, USA, {\tt\small
%      \{rvega7\}@gmu.edu}}}
%      
\maketitle 

\begin{abstract}
%After decades of swarms research, the connections between local interactions of agents in a system and the induced emergent collective behavior are still unclear. Even if a specific interaction rule is known to produce some particular emergent behavior, the conditions under which this works may still not be well understood. By thinking of the agents similar to how chemists might think of water molecules in different macrostates, we take a less engineered and more scientific approach in which we aim to characterize the conditions under which a particular emergent behavior appears or not, rather than designing the local control systems to force the desired behavior. This allows us to study conditions under which things happen  which can then indirectly be used to produce desired emergent behaviors or properties by understanding the exact conditions under which they self-organize that way. 
 Emergence and emergent behaviors are often defined as cases where changes in local interactions between agents at a lower level effectively changes what occurs in the higher level of the system (i.e., the whole swarm) and its properties. However, the manner in which these collective emergent behaviors \textit{self-organize} is less understood. 
 %This paper for the first time attempts to leverage the knowledge between chemistry, fluid mechanics, and robot swarms. By forming these connections, we attempt to apply established methodologies and tools from these these domains to uncover how we can better comprehend swarms, specifically when looking at the emergent behaviors produced in a swarm. 
 The focus of this paper is in presenting a new framework for characterizing the conditions that lead to different macrostates and how to predict/analyze their macroscopic properties, allowing us to indirectly engineer the same behaviors from the bottom up by tuning their {environmental} conditions rather than local interaction rules. We then apply this framework to a simple system of
 binary sensing and acting agents as an example %that uses a binary controller that has been shown to create the milling behavior and provide a set that constrains the parameters of the system depending on the desired milling radius.
 to see if a re-framing of this swarms problem can help us push the state of the art forward.  By first creating some working definitions of macrostates in a particular swarm system, we show how agent-based modeling may be combined with control theory to enable a generalized understanding of controllable emergent processes without needing to simulate everything. Whereas phase diagrams can generally only be created through Monte Carlo simulations or sweeping through ranges of parameters in a simulator, we develop closed-form functions that can immediately produce them revealing an infinite set of swarm parameter combinations that can lead to a specifically chosen self-organized behavior. 
 While the exact methods are still under development, we believe simply laying out a potential path towards solutions that have evaded our traditional methods using a novel method is worth considering. Our results are characterized through both simulations and real experiments on ground robots.
\end{abstract}

\section{Introduction}\label{se:intro}

%\margin{just bring in citations and explain the citation most relevant to our work here. }

%\margin{mention (with citations) how others have used certain metrics to do clustering and identify different emergent behaviors and we want to take this further and extend it by introducing new metrics that are directly tied to different notions of usefulness or even problem definitions}

% \db{I wonder if the plots (or at least some of them would be better placed later when you actually talk about the corresponding swarm experiments. The intro is interesting, but I'm worried it goes too much into the weeds about chemistry without really saying anything about what the paper results are. }

Swarms have been purported to be useful in many real world applications including pollution monitoring \cite{GZ-GKF-DPG:11}, disaster management systems \cite{HK-CWF-IT-BS-ER-KP-AW-JW-12}, surveillance \cite{MS-JC-LP-JT-GL-AT-VV-VK:14}, and search and rescue \cite{RA-JJ-BA-EM:20}; but after 50+ years of research we still do not see any engineered swarms solving practical problems or providing any real benefits to society. This may be due to the fact that we are yet to fully understand how we can control swarms of robots due to the naturally complex interplay between agents that leads to emergent behaviors~\cite{JT:05, JT:05ten, OTH:07}.

% Even if a specific interaction rule is known to produce some particular emergent behavior, the conditions under which this works may still not be well understood~\cite{JT:05, JT:05ten, OTH:07}. By thinking of the agents similar to how chemists might think of water molecules in different macrostates, we take a less engineered and more scientific approach in which we aim to characterize the conditions under which a particular emergent behavior appears rather than designing the local control systems to force the desired behavior. By studying and understanding the exact conditions under which things occur, we can then use the same knowledge to produce desired emergent behaviors with desired properties.

The classic approach to \textit{engineering} robot swarms is to use top-down methods with a specific macro objective or metric in mind that the multi-agent system is designed to optimize~\cite{AO-MG-AK-MDH-RG:19, MG-JC-WL-TJD-RG:14, DS-CP-GB:18, MG-JC-WL-TJD-RG:14clust}. This ignores principles of self-organization and instead aims to engineer very specific individual agents and carefully control their local interactions to guarantee certain macro-behaviors `emerge.' For example, there are many collective swarm behaviors that have been well studied including flocking, shepherding, shape formation, and collective transport; however, the agents used to create those behaviors are generally sophisticated and often require sensing/actuating/communicating capabilities that are not pragmatic~\cite{ALA-MGCAC-NDF-ML-GV:18, YC-YRH-MRD-ALB:07, AO-MG-RG:17, YM-HG-YJ:13, MR-AC-JW-GH-JM-RN:13}. Instead, we propose a more \textit{scientific} approach in which we aim to understand what types of environments and interactions naturally tend to self-organize into different macrostates through emergence, similar to how it's done in chemistry.  

\begin{figure}[h!]
    \centering
    {\includegraphics[width=.85\linewidth]{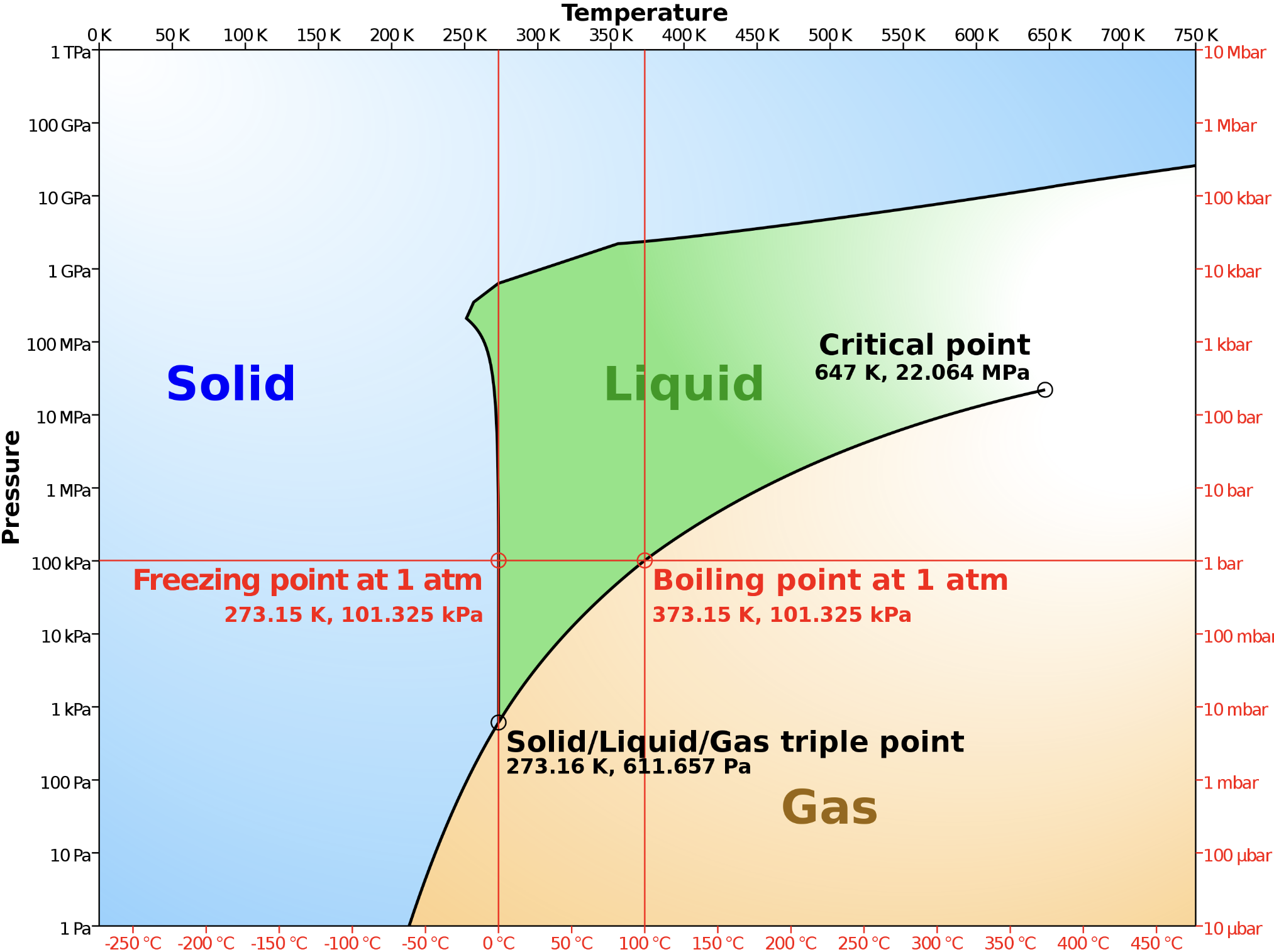}}
    \caption{Phase diagrams of the macrostates of matter of $H_20$ molecules~\cite{wiki_phase}.}
    \label{fig:water_phases}
\end{figure}

Connections between swarms and chemistry are not entirely new: the idea of `swarm chemistry'  was first coined by Sayama in 2009~\cite{HS:09} where it was found that mixing agents with different control rules led to fascinating new behaviors. However, these swarm chemistry ideas only consider combining two or more different `elements' or `species' together~\cite{HS:09, HS:11, HS:12}. We believe the connection to chemistry goes much deeper and exploring these ideas for even homogeneous agents is a missed opportunity. Furthermore, we can borrow the use of phase diagrams from chemistry as a key tool to visualize the phases of the swarm and use them as look-up tables for how to deploy swarms with a particular macrostate the same way the phase diagram of water(Fig.~\ref{fig:water_phases}) is used to determine what temperature and pressure is needed to make solid ice~\cite{BP-MH-MJP:13}. The only complication is that we are dealing with a much higher dimensional parameter space as we are dealing with sensing and acting agents rather than inanimate water molecules. The question remains fundamentally the same: can we identify and visualize the subsets of a swarm parameter space to help us know immediately how many swarming robots are needed for a desired behavior to self-organize without running new simulations every time? 

%However, these earlier works treat the agents as particles and use dynamic rules such as the Boid's algorithm \cite{CWR:87} or similar attractive/repulsive potentials.

% {\color{blue}
%need to mention radhika/gauchi/etc on the milling problem. this is the perfect example to test our framework because the local interaction rule they found has been verified several times in both simulation and hardware -- but still none of their papers have been able to provide answers on how exactly to make the mill without trial and error. we are the first to get rid of any trial and error for the specific milling problem.

Here, we use milling as the main example as it a rather simple emergent behavior that has been shown to occur using different local interaction rules and has been verified multiple times in both simulation and hardware~\cite{AC-CKH:18, MG-JC-TJD-RG:14, DB-RT-OH-SL:18, DS-CP-GB:18,FB-MG-RN:21}. However, there are still no concrete answers that guarantees a mill can be created without the use of trial and error. The work~\cite{FB-MG-RN:21} lays the groundwork for obtaining some theoretical limits on the size of a mill that can be formed utilizing the same simple binary sensing-to-action method first discovered in~\cite{gauci2014evolving} (although with slightly different sensing than our agents), and proposes an equation and conditions under which the emergent milling radius can be predicted. As far as the authors are aware, this is currently the best answer in the literature to understanding the milling behavior but we have been unable to independently verify their equation in most cases and propose a new equation that we are able to confirm through geometric proof and simulation.

%When the goal of an engineer is a particular emergent behavior, it becomes tempting and easy to simply endow individual agents with enough capabilities to ensure the global behavior can be achieved. Unfortunately this approach does not help us understand how exactly a swarm of agents at a certain capability level can exhibit different emergent behaviors. Going back to chemistry and water, water molecules have no agency, they cannot sense their environment or take actions; but, by controlling their environments and inducing changes in the macroscopic properties of the whole system, we can get them to do a wide assortment of things.

%In order to properly study how low-level interactions manifest into group behaviors, we propose that complicated robots makes studying swarms much harder. If even zero-agency water molecules can exhibit different macro-behaviors (e.g., ice, water, vapor) \cite{BP-MH-MJP:13}, what can a swarm of very simple robots do together? Can simple robots exhibit different `states of matter' depending on their `temperature' and `pressure'? 

\paragraph*{Statement of Contributions}
This paper presents a novel framework for thinking about robot swarms by drawing connections to chemistry and fluid mechanics. Our contributions are threefold. First, we show how small differences in the environmental configurations or inter-agent interaction rules can lead to fundamentally different macrostates, as well as present an example of a set of conditions that guarantees the milling circle behavior. Second, we formalize a brand new connection between swarms and fluid mechanics to seek empirical relationships between the properties of a swarm in a particular macrostate and the agent-level states. Finally, we validate all results in simulation and a few with real robot experiments.

\newcommand{\omegamax}{\omega_\text{max}}

\section{Problem Formulation}\label{se:problem_formulation}
With the goal of formalizing a new framework for swarm analysis, we consider a very simple swarm of~$N$ self-propelled Dubins' vehicles moving in a 2D environment~$D \subset \real^2$. The 2D position and orientation of each robot at time $t$ is given by $\mathbf{p}_i(t) = [x_i(t),y_i(t)]^T \in D$ and~$\theta_i(t) \in [-\pi,\pi)$, respectively, with %kinematics
\begin{align}\label{eq:real_func}
    \left[ \begin{array}{c} \dot{x}_{i}(t) \\ \dot{y}_{i}(t)\\ \dot{\theta}_{i}(t)\end{array} \right]  = \left[ \begin{array}{c} v\cos \theta_{i}(t) \\ v \sin \theta_{i}(t) \\ u_i(t)  \end{array} \right].
\end{align}

\begin{figure}[t]
    \centering
    \includegraphics[width=1\linewidth]{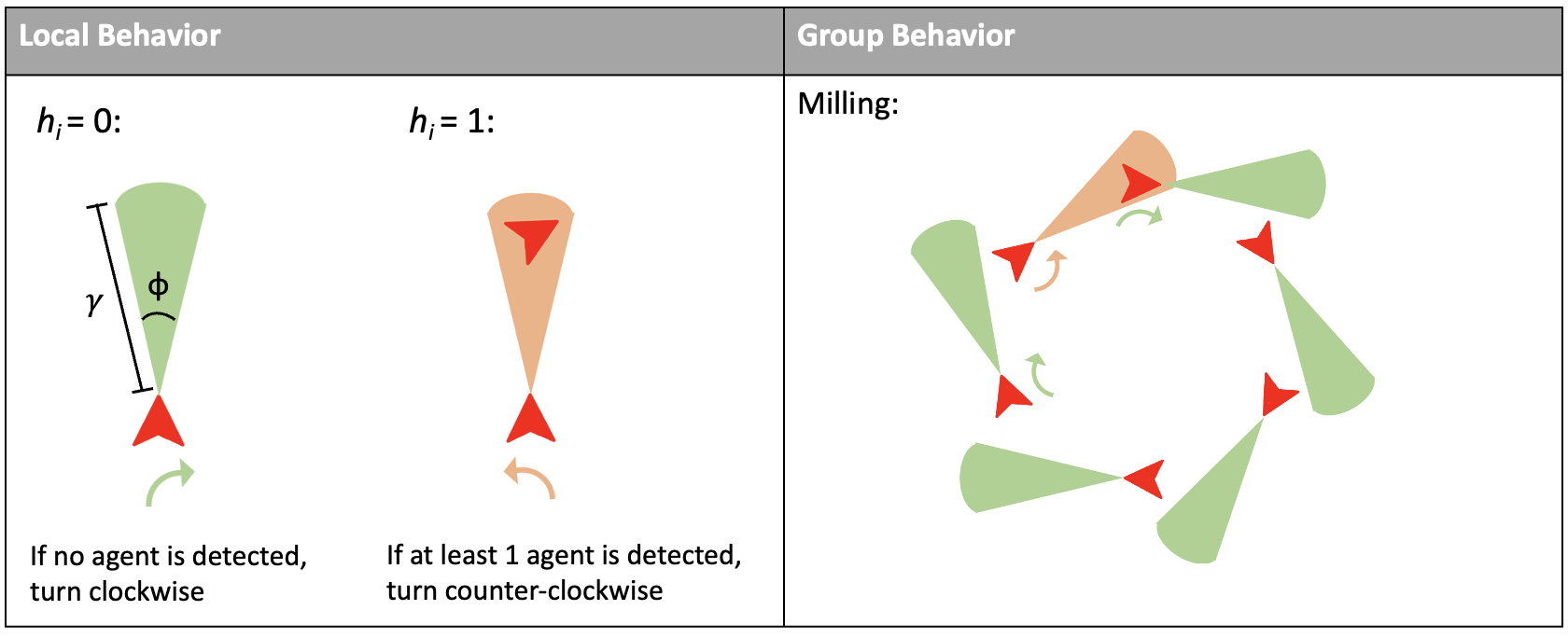}
    \caption{Different local interaction rules leading to different emergent group behaviors (under certain conditions).}
    \label{fig:milling_figure}
\end{figure}

Our agents only have a binary sensor that is triggered when at least one other agent is within the sensor's field of view, which is the conical area in front of the robot with range~$\gamma>0$ and opening angle~$\phi>0$ as shown in  Fig.~\ref{fig:milling_figure} and denoted as $\operatorname{FOV}_i$ for agent $i$. We assume that all agents are always moving with a constant velocity~$v > 0$ and that their turning rate~$u_i$(t) is a function of their sensor output $h_i(t)$:

{
\begin{align}\label{eq:output}
    h_i(t) = \begin{cases}
                 1 & \text{if } \exists j \neq i, s.t. ~ p_j \in \operatorname{FOV}_i , \\
                 0 & \text{otherwise.}
             \end{cases}
\end{align}
}

Now, in order to define a macrostate of a swarm, we look towards statistical mechanics and thermodynamics where macrostate is often defined as being a set of microstates that share specific macroscopic properties \cite{CRS-CM:03}. These macroscopic properties (commonly temperature, pressure, density, etc.) are essentially properties that are only applicable to the system as a whole. Here we use the term microstates to refer to the collective states of all the agents. More formally, we define the full state of each agent at time $t$ as $\mathbf{z}_i(t) = [\mathbf{p}_i(t),\theta_i(t)]^T$ and the microstate of the swarm as the position of all the agents $Z(t) = [z_i(t), ..., z_N(t)]$. 

Our goal here is \textbf{not} to design a specific control input to exhibit some desired behavior; rather we attempt to identify/characterize the conditions that lead to multiple different macro-behaviors and the macroscopic properties applicable therein. Therefore, we consider a very simple binary sensing-to-action controller that has already been discovered to exhibit interesting emergent behaviors,
\begin{align}\label{eq:control}
   u_i(t) = \begin{cases}
                 \omega & \text{if } h_i(t) = 1 ,\\
                 -\omega & \text{otherwise,}
             \end{cases}
\end{align}
where~$\omega > 0$ is the selected turning rate of all the agents. This simple binary controller was chosen as it was found to be capable of producing the emergent behavior where the agents form a rotating circle (or milling)~\cite{gauci2014evolving, DB-RT-OH-SL:18, DS-CP-GB:18, FB-MG-RN:21}, see Fig.~\ref{fig:milling_figure}. However, this milling macrostate only appears under certain nontrivial conditions that have not yet been concretely established.

A system of agents with the controller \eqref{eq:control} may exhibit the milling circle behavior (macrostate) but it is not guaranteed and is dependent on not only the initial positions of the agents but also of the various parameters that have an influence on the interactions between agents. We denote $\PP$ as the parameter space~$\real_{+}^3 \times [0, 2\pi) \times \mathbb{N}$, containing the agent parameters $\gamma, v,\omega,$ and $\phi$ and the environment parameter $N$. The size of the environment and presence of obstacles would undoubtedly affect the swarm but for now we assume the system to be in an open environment and therefore can disregard these for now.

We can then formally define the problems this paper aims to answer.
% \begin{problem}[Predicting the Macrostate of the Swarm]\label{pr:chem}
%     {\rm
%     Given the kinematics~\eqref{eq:real_func} with direct sensing-to-action controller~\eqref{eq:control}, if $Z(0)$ and $\PP$ are known, can we predict the future macrostate of the system?
%     }
% \end{problem}

\begin{problem}\longthmtitle{Characterizing the Sufficient Conditions of the Milling Macrostate}\label{pr:chem}
    {\rm
    Given the kinematics~\eqref{eq:real_func} with direct sensing-to-action controller~\eqref{eq:control}, under what initial conditions $Z(0)$ and parameters $\PP$ will the milling macrostate be guaranteed to emerge?
    }
\end{problem}

% {\color{red}
% \begin{problem}[Analyzing Properties within the Macrostates]\label{pr:mech}
% {\rm
% When the initial conditions $Z(0)$ and parameters~$\PP$ give rise to the milling macrostate, what are the properties applicable to that specific macrostate?
% }
% \end{problem}

\begin{problem}[Predicting the Properties of Milling]\label{pr:mech}
{\rm
When the initial conditions $Z(0)$ and parameters~$\PP$ give rise to the milling macrostate, how can we determine what the properties of the milling swarm will be (e.g. radius of milling circle)?
}
\end{problem}

\section{Swarm Chemistry}\label{se:swarm_chem2}

\subsection{Identifying Possible Macrostates}
We first begin by defining the various high-level properties that can help determine the macrostates this particular system is capable of converging to.

\subsubsection{Macroscopic Properties}
Just as thermodynamics uses the well established notions of temperature, pressure, energy, density, and more to group microstates of a molecular system into macrostates, we looked at multiple measurable properties (metrics) to identify exactly how the behaviors of the swarm may differ.  Although, \cite{HH-KJ-JL:21} identified and defined the macroscopic properties they called `swarm temperature', `swarm pressure', and `swarm density'; their system was based on attractive/repulsive agent behaviors similar to how real molecules interact with one another and therefore isn't directly applicable for our more simplistic model. Instead various other metrics are selected since they were thought to have a good chance in distinguishing between macrostates, although it should be noted that these are not necessarily the only viable options that could have been used.

\begin{table}[t]
\begin{center}
  \begin{tabular}{|l|c|p{4.1cm}|}
    \hline \hline
    Name & Variable & Equation \\ \hline
    Average  Speed & $\overline{V}$ & $ \frac{1}{N} \sum_{i=1}^{N} ||\dot{\mathbf{p}}_i||_2 $ \\ \hline 
    Group  Rotation & $\overline{\Omega}$ & $ \sum_{i=1}^{N} (\dot{\mathbf{p}}_i \times \frac{\mathbf{p}_i - \mu}{||\mathbf{p}_i - \mu||} )$ \\ \hline
    Ang.  Momentum & $\overline{L}$ & $\frac{1}{R \cdot N} \sum_{i=1}^{N} (\dot{\mathbf{p}}_i \times (\mathbf{p}_i - \mu))$ \vspace*{1ex}\\ \hline
    Scatter & $\overline{s}$ & $\frac{1}{R^2 \cdot N} \sum_{i=1}^{ N} ||\mathbf{p}_i - \mu||^2$ \vspace*{1ex} \\ \hline
    Radial  Variance & $\overline{\xi}$ & $\frac{ \sum_{i=1}^{ N} (||\mathbf{p}_i - \mu|| - \frac{1}{N} \sum_{i=1}^{N} ||\mathbf{p}_i - \mu||)^2}{{R^2 \cdot N}}$\vspace*{1ex} \\ \hline
    Circliness & $\overline{c}$ & $\frac{\max_{i \in N} ||\mathbf{p}_i-\mu|| - \min_{i \in N}||\mathbf{p}_i-\mu||}{\min_{i \in N}||\mathbf{p}_i-\mu||}$ \vspace*{1ex} \\ \hline
    %Minimum Separation & $\overline{w}$ & $$\min_{i,j \in N}||\mathbf{\xi}_i - \mathbf{\xi}_j||$$\\ \hline
  \end{tabular}
\end{center}
  \caption{Measurable properties of the swarm measurable by an external observer }
  \label{tab:properties}
  \end{table}

\begin{figure}[t]
\centering
  \includegraphics[width=1\linewidth]{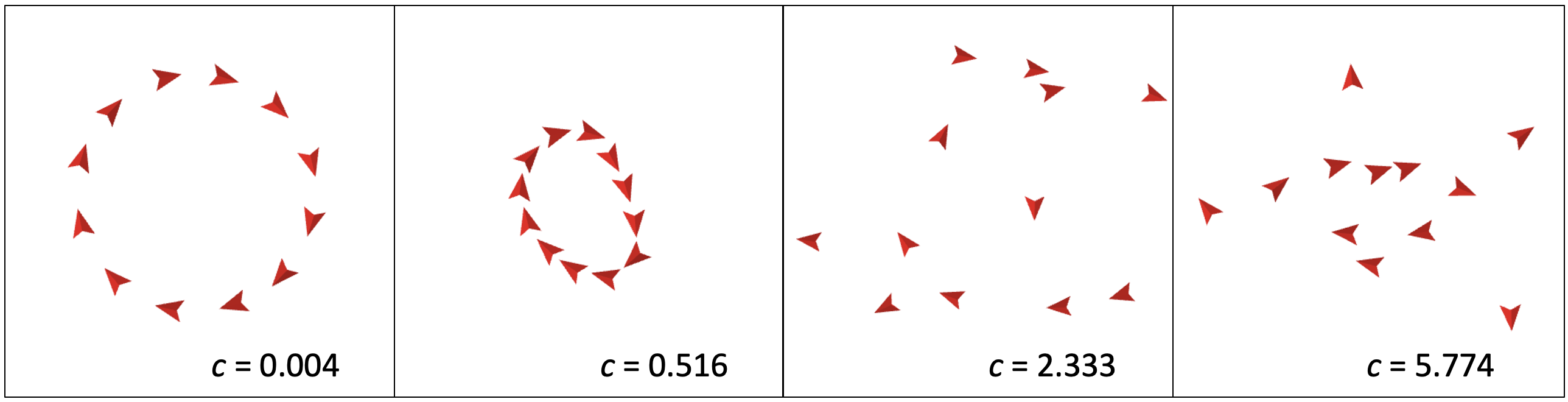}
  \put(-214,7){\tiny$-$}
  \put(-153,7){\tiny$-$}
  \put(-92,7){\tiny$-$}
  \put(-33,7){\tiny$-$}
  \caption{Examples of different values of $\overline{c}$.}
  \label{fig:circliness_plots}
\end{figure}

The first five properties in Table~\ref{tab:properties} are the same as used in \cite{DB-RT-OH-SL:18} where  a novelty search algorithm  was used to discover behaviors possible given a specific capability set (i.e. $\PP$ remained constant and instead the controller $u_i$ was varied). The milling behavior was one of the behaviors found when the simulations were clustered by these properties. In addition to these, since milling was known to be possible, we also chose to measure the ``circliness" of the system, similar to the metrics in \cite{CT-CL-CN:20}, that essentially looked at the distance of the closest and farthest agent from the center of mass $\mu(t) = \frac{1}{N}\sum^{N}_{i=1}\mathbf{p}_i(t)$, where~$\overline{c}=0$ represents a perfect mill. Fig~\ref{fig:circliness_plots} shows snapshots of what different values of~$\overline{c}$ look like.  

%Letting~$r_\text{max} = \max_{i \in N} ||\mathbf{p}_i-\mu||$ and~$r_\text{min} = \min_{i \in N}||\mathbf{p}_i-\mu||$, this is
%\begin{align}\label{eq:circly}
% \overline{c} = \frac{\max_{i \in N} ||\mathbf{p}_i-\mu|| - \min_{i \in N}||\mathbf{p}_i-\mu||}{\min_{i \in N}||\mathbf{p}_i-\mu||} = \frac{r_\text{max}}{r_\text{min}} - 1 ,
%\end{align}

\begin{figure}[b]
\centering
  \subfigure[]{\includegraphics[width=.35\linewidth]{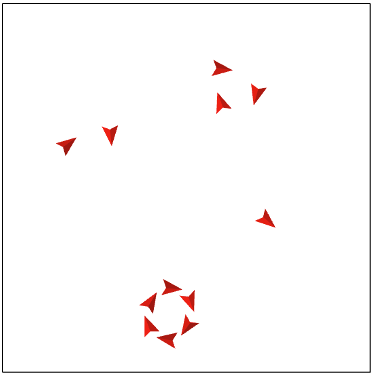}} \hspace*{6ex}
  \subfigure[]{\includegraphics[width=.35\linewidth]{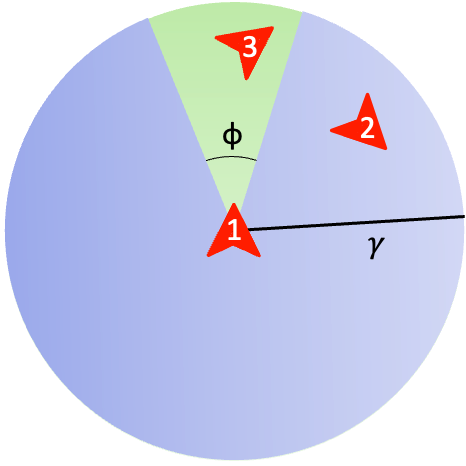}}
  \caption{Simulation showing (a) when the system has broken apart into multiple groups, and (b) zooming in on the 3 agents at the top right to view them as a single sub-swarm.}
  \label{fig:multi_group12}
\end{figure}

%\paragraph{When is One Swarm Really Two Swarms?}
The metrics above always assumes the entire group of agents should be considered as part of a single swarm system. Here we formalize when exactly it makes sense to partition our group of~$N$ agents into two or more distinct `swarms'. To address this we turn to graph theory and in particular the algebraic connectivity of the interaction graph of the agents. Fig~\ref{fig:multi_group12}(a) shows a scenario in which a human may quickly partition the~$N = 12$ agents in 4 distinct `groups' or sub-swarms, even if one of the groups is degenerate with only a single agent. 

%These properties above would be attainable by an outside observer of the system analyzing the trajectories of all the agents. Additionally, we also measured the algebraic connectivity of the system which conversely requires information that would only be available if the observer was within the system or had access to the all the information of the agents. This additional property was chosen after seeing some behaviors in simulation where the system would break apart into multiple groups, Fig~\ref{fig:multi_group12}(a).

Algebraic connectivity uses graph theory to measure how connected the system is and we review here only the most pertinent basic information needed but refer readers not familiar with graph theory to~\cite{FB-JC-SM:09}. 
The \emph{adjacency matrix} of a digraph $\mathcal{G}\triangleq\left(\mathcal{V},\mathcal{E}\right)$ consisting of vertices~$\mathcal{V}$ and ordered edges~$\mathcal{E}\subseteq \VV \times \VV$ is
denoted by $A_{\mathcal{G}}=[a_{ij}]$, which is an $N \times N$ matrix defined entry-wise as $a_{ij}=1$ if edge $(v_{j},v_{i})\in\mathcal{E}$, and $a_{ij}=0$ otherwise. The matrix $A_\GG$ is \emph{irreducible} if and only if its associated graph $\mathcal{G}$ is strongly connected. The Laplacian is given by~$L = D - A$ and the algebraic connectivity is its second smallest eigenvalue~$\lambda_2(L)$. It is well known that when~$\lambda_2 > 0$, the graph~$\GG$ is connected. 

Here we begin by constructing the range-limited visibility graph $\mathcal{G}_\text{vis-disk}$~from\cite{FB-JC-SM:09} by putting the edge~$(v_j,v_i) \in \edges$ if and only if agent~$j$ is being seen by agent~$i$, i.e., $p_j \in \FOV_i$. Note that in general this is a dynamic (state-dependent) directed graph and may result in an adjacency matrix that is time-varying and not symmetric. To create an easily verifiable sufficient condition that tells us when the graph is disconnected, we construct a larger but simpler r-disk graph $\mathcal{G}_\text{disk}$, where the edge~$(v_j,v_i) \in \edges$ if and only if agent~$j$ is within $\gamma$ of agent~$i$, i.e., $||p_j - \mathbf{p}_i|| < \gamma$. Then, if~$\mathcal{G}_\text{disk}=0$, according to Lemma~\ref{le:subgraph} below the system is no longer a single swarm.
%$N = 12; \omega = 30; v = 0.15$  -- FOR CLEAR PHASES OF VISION CONE VS DISTANCE DIAGRAM
\begin{lemma}\longthmtitle{Sufficient condition for multi-swarm analysis}\label{le:subgraph}
If~$\lambda_2(L_\text{disk}) = 0$, then~$\lambda_2(L_\text{vis-disk}) = 0.$
\end{lemma}

\begin{example}[Connection between~$\GG_\text{disk}$ and~$\GG_\text{vis-disk}$]
{\rm Given the state shown in Fig.~\ref{fig:multi_group12} (b) with~$N=3$, we can construct the adjacency matrices for both $\mathcal{G}_\text{vis-disk}$ and $\mathcal{G}_\text{disk}$,
\begin{align*}
A_{\mathcal{G_\text{vis-disk}}}= \begin{bmatrix} 0 & 0 & 1 \\ 0 & 0 & 0 \\ 0 & 0 & 0 \end{bmatrix}, ~~~~A_{\mathcal{G_\text{disk}}}= \begin{bmatrix} 0 & 1 & 1 \\ 1 & 0 & 1 \\ 1 & 1 & 0 \end{bmatrix}.
\end{align*}

From this, it is straightforward to see that $\mathcal{G}_\text{vis-disk} \subset \mathcal{G}_\text{disk}$, and if the larger graph~$\GG_\text{disk}$ is not connected, then~$\GG_\text{vis-disk}$ must be not connected also. 
\oprocend
}
\end{example}

\subsubsection{Macrostates}
Using these metrics, we can now attempt to define the constraints of the observed behaviors/macrostates in Table~\ref{tbl:macrostates1}. These constraints were chosen rather subjectively after analyzing the simulated systems. For example, the values $c_1$ and $c_2$ represent the boundary of the circliness metric between $M/E$ and $E/U$, respectively, and were chosen based on the appearance of the system, examples of what these values represent can be seen in Fig~\ref{fig:circliness_plots}. Simulations of the system were run in Netlogo \cite{UW:99}, were the time-steps was reduced small enough such that reducing it further had no change in behavior in order to emulate the continuous-time model~\eqref{eq:real_func} used. It should also be noted that all the simulations were initially setup such that r-disk graph $\mathcal{G}_\text{disk}$ was strongly connected.

         \begin{table} [t]
        \centering
        \begin{tabular}{|c|c|p{4cm}|}
            \hline
             \hspace*{-1ex}Macrostate\hspace*{-1ex} & Name  & Constraints \\
            \hline \hline
            $U$ & Uncharacterized &  $\lim\limits_{t\rightarrow \infty}\overline{c}(t) \geq c_2,  \newline \liminf\limits_{t\rightarrow \infty} \lambda_2(L_\text{disk}(t)) > 0 $ \\ \hline
            $E$ & Ellipsoidal&  $ c_1 \leq \lim\limits_{t\rightarrow \infty}\overline{c}(t) < c_2, \newline \liminf\limits_{t\rightarrow \infty} \lambda_2(L_\text{disk}(t)) > 0 $ \\ \hline
            $M$ & Milling& $\limsup\limits_{t\rightarrow \infty}\overline{c}(t) < c_1, \newline \liminf\limits_{t\rightarrow \infty} \lambda_2(L_\text{disk}(t)) > 0 $ \\  \hline
            $P$ & Pulsing Mill& $\limsup\limits_{t\rightarrow \infty}\overline{c}(t)<c_1$, \newline $\limsup\limits_{t\rightarrow \infty} \overline{\Omega}(t) - \liminf\limits_{t\rightarrow \infty} \overline{\Omega}(t) > 0, \newline \liminf\limits_{t\rightarrow \infty} \lambda_2(L_\text{disk}(t)) > 0$ \\  \hline
            $C$ & Collapsing Circle & $\limsup\limits_{t\rightarrow \infty} \overline{c}(t) - \liminf\limits_{t\rightarrow \infty} \overline{c}(t) > 0$\\ \hline
            $S$ & Separated Groups & $  \limsup\limits_{t\rightarrow \infty}\lambda_2(L_\text{disk}(t)) = 0$\\ \hline
            %D$ & Dispersed & $  \liminf\limits_{t\rightarrow \infty} \overline{w}(t) > 2(\frac{v}{\omega})$\\ \hline

        \end{tabular}
        \caption{Table of Macrostate Definitions}
        \label{tbl:macrostates1}
    \end{table}

After running several parameter sweeps in $\PP$, we found that systems running the same binary controller may  produce various different behaviors depending on the parameters. More specifically we found that a system running the binary controller \eqref{eq:control} can produce six behaviors: a well-shaped and stable milling circle ($M$), a pulsating circle ($P$), an imperfect ellipsoidal formation ($E)$, an unorganized group motion ($U$), a constantly reforming and collapsing circle ($C$), and behavior where the system breaks into multiple sub-groups ($S$).

\subsection{Swarm Macrostate Phase Diagrams}\label{se:phase_diagrams}
The aim of this section is to provide tools to enable the characterization of emergent behaviors from a swarm of agents. As chemists use phase diagrams to predict the state of matter of chemical substances at certain conditions, we develop similar phase diagrams that visualizes how different conditions lead to different macrostates for this particular simple swarm. However, unlike the phase diagrams used in chemistry, we cannot directly use the macroscopic properties of this system as the axes of the diagrams. Although still a very simple system of agents, the additional agency added with the binary sensor-to-actuator control amplifies the complexity when analyzing the swarm. Instead, we look at the parameters that may affect the measured macroscopic properties and use those to create phase diagrams.

\begin{figure}[t]
    \centering
    \includegraphics[width=8cm]{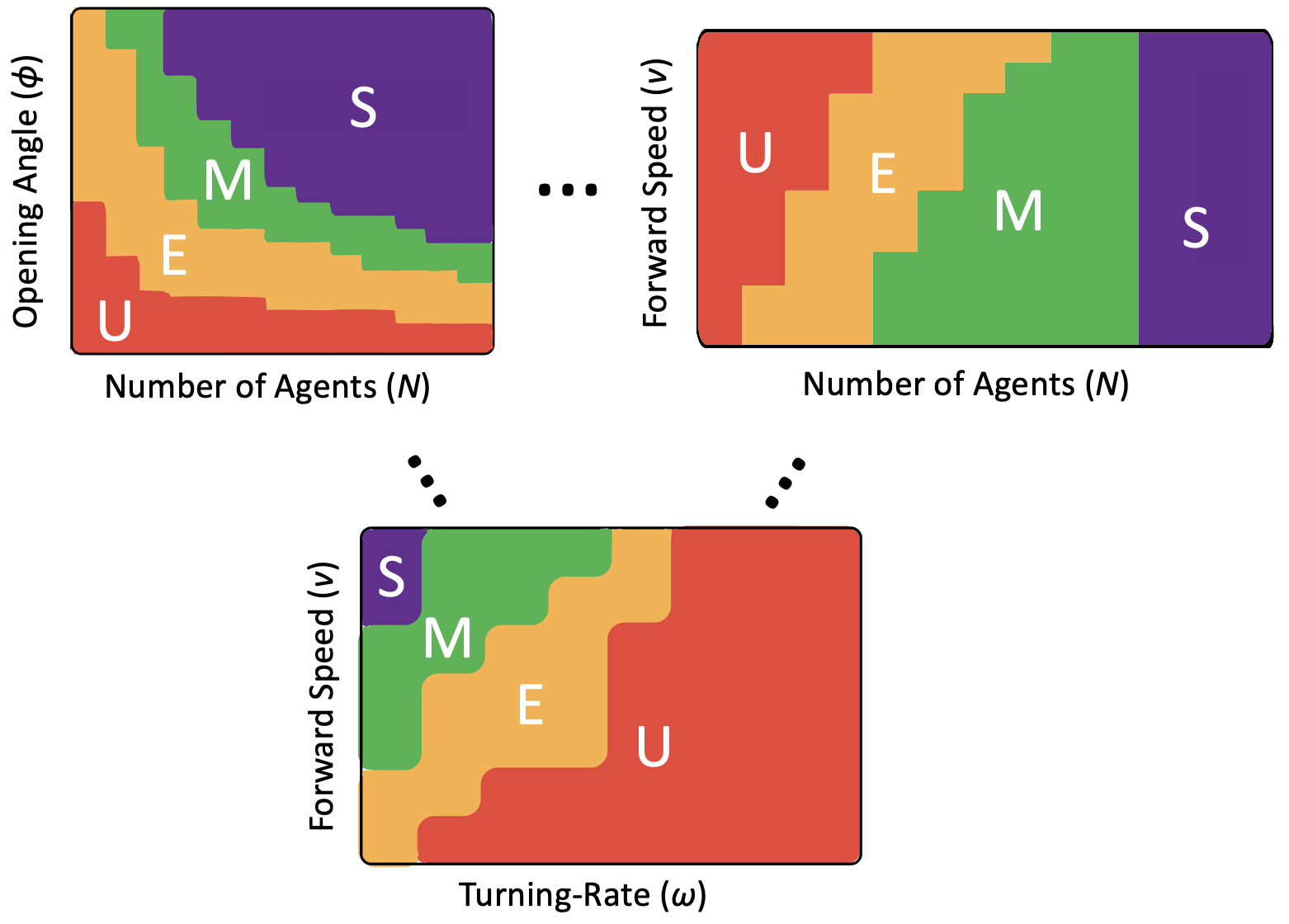}
    \caption{Multiple phase diagrams showing how various conditions can lead to different behaviors. This shows that how these diagrams are only 2D slices of a much more complicated and higher dimensional space. 
    }
    \label{fig:multiple_phase_diagrams}
\end{figure}

In this particular system, we consider the parameters $\PP$ to be the independent variables that can be changed (e.g. the number of agents $N$, the forward speed of the agents~$v$, turning-rate $\omega$, the vision distance $\gamma$, and opening angle of the FOV $\phi$). Unfortunately, sweeping through all these parameters separately results in a 5-dimensional phase space that is not quite pragmatic (Fig~\ref{fig:multiple_phase_diagrams}); each phase diagram here is only a 2D slice of this space and it is difficult to picture this larger space by just looking at them a few at a time. This shows how much larger the full search space is for this quite simple swarm and the difficulty of efficiently predicting when these behaviors occur. Furthermore, it can be seen that some of the introduced macrostates are not present in these phase diagrams (i.e. $P$ and $C$), this is due to the instability of these behaviors that limits them to occur in very specific initializations, whereas the simulated runs only initialized the agents such that their interaction graph was strongly connected.

However, these diagrams do give us better insight of where we can begin looking for the conditions where milling is sure to occur. For example, while keeping everything else constant, the phase diagram of $\phi$ vs $N$ (Fig.~\ref{fig:multiple_phase_diagrams}) shows a thin band where the milling macrostate appears, hinting at a stronger relationship between $N$ and $\phi$ than the other parameters.

\section{Swarm Mechanics}\label{se:swarm_mechanics}

\begin{figure}[t]
    \centering
    \includegraphics[width=8cm]{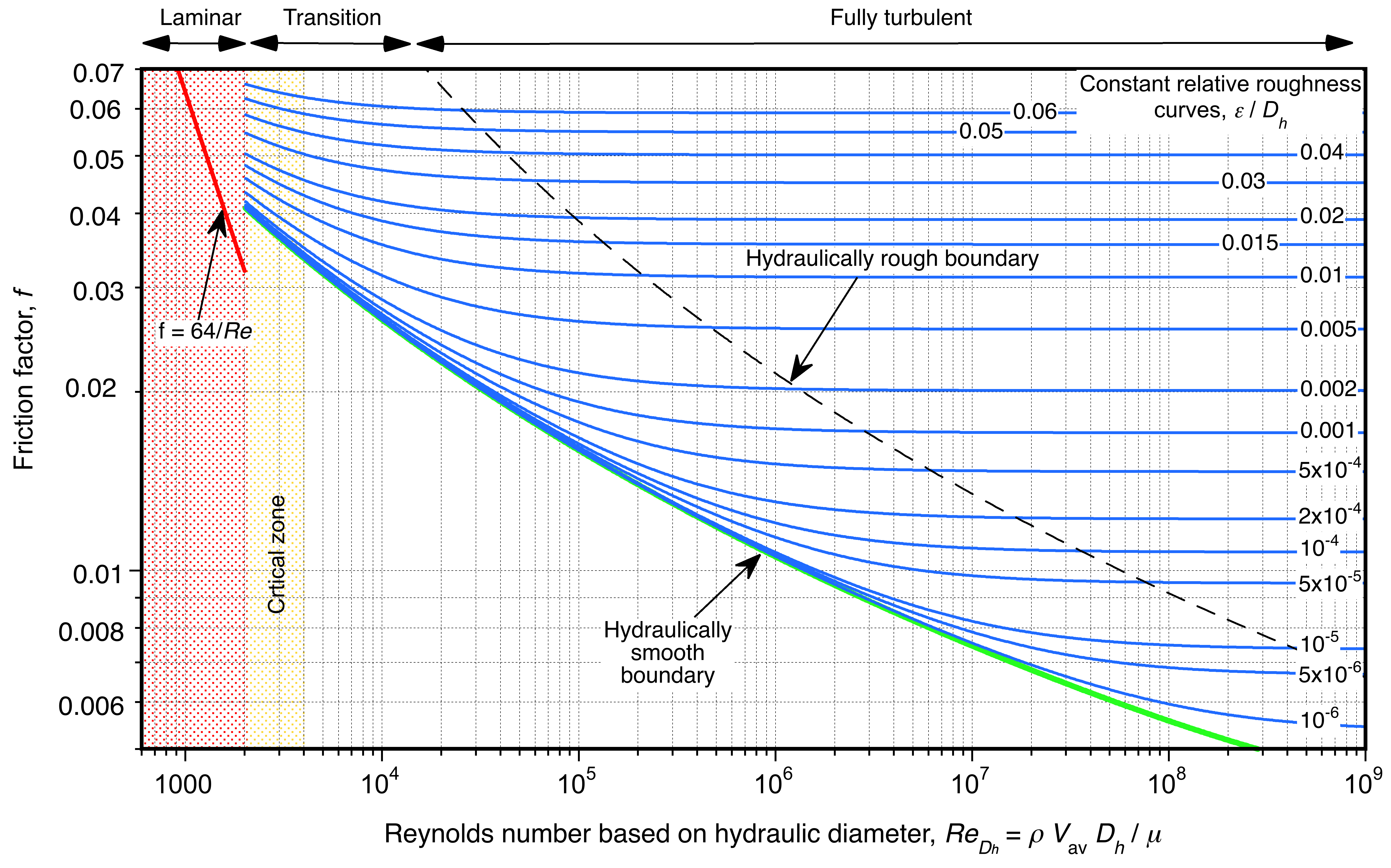}
    \caption{Moody diagram showing the transition from laminar to turbulent flow based on the Reynolds's number of various relative roughness~\cite{LJG:22}}
    \label{fig:moody-diagram}
\end{figure}

The above methods we call Swarm Chemistry have focused on solving Problem~\ref{pr:chem} by determining what macrostate the agents of a swarm are. Although not entirely characterized, we have found conditions in simulation in which the agents self-organize into the milling phase or other macrostates. In this section we want to further analyze a single state of matter (e.g., milling) and find empirical relations between the conditions of the swarm and its macrostate properties.
Similar to how the Darcy friction factor of flowing fluid is a function of its Reynolds number in Fig.~\ref{fig:moody-diagram}, can we solve Problem~\ref{pr:mech} by finding functions and conditions that predict the radius of the emergent circle~$R_m$ in the milling phase?

We begin our solutions to Problems~\ref{pr:chem} and~\ref{pr:mech} by analyzing the geometry of a mill (similar to \cite{FB-MG-RN:21}) and stating the sufficient conditions to guarantee the preservation of a perfect mill in Proposition~\ref{prop:at_manifold}.

\begin{proposition}\longthmtitle{Maintaining the Perfect Mill}\label{prop:at_manifold}
The subspace of the parameter space~$P$ that guarantees the possibility of a mill is given by
  %Given $Z(0)$ and $P$ such that the agents begin in a circle, $\lambda_2 (L_\text{disk}(0)) > 0$, and 
    \begin{align*}
   \PP^* & \triangleq \{P \in \mathcal{P} | \phi = \frac{2\pi}{N}, \frac{v}{\omega} \leq \frac{\gamma}{2\sin(\pi/N)}\}.
\end{align*}
More specifically, for any~$P \in \PP^*$, there exists~$Z(0)$ such that~$\overline{c} = 0$ is an invariant property. Moreover, the emergent milling radius will be
    \begin{align}\label{eq:R_m}
    R_m = \frac{\gamma}{2\sin(\frac{\pi}{N})}.  
    \end{align}
\end{proposition}

% \begin{align*}
%     P \in \PP^* & \triangleq \{P \in \mathcal{P} | \phi = \frac{2\pi}{N}, \frac{v}{\omega} \leq 5, \gamma = 10\sin(\frac{\pi}{N}) \},
% \end{align*}

See appendix for proof. \oprocend

% \begin{align}\label{eq:Rm}
% R_m = \frac{\rho}{cos\frac{\phi}{2}-cos(\frac{2\pi}{N} - \frac{\phi}{2})}.
% \end{align} 

\begin{figure}[t]
\centering
  \includegraphics[width=.5\linewidth]{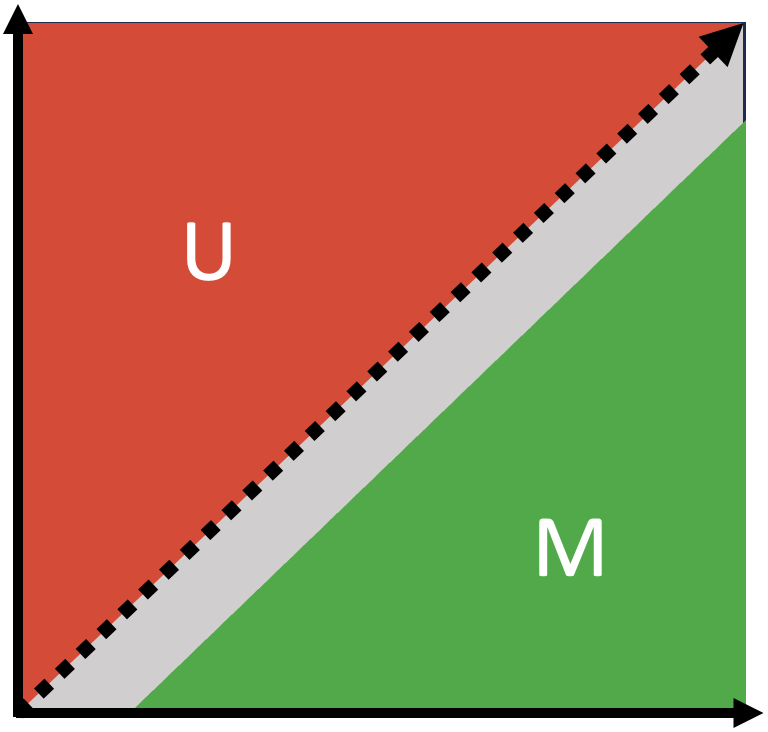}
  \put(-135,65){$\frac{v}{\omega}$}
  \put(-70,-10){$R_m$}
  \caption{Turning radius ($\frac{v}{\omega}$) vs $R_m$ showing an almost clean separation and relationship between the two values and the macrostate produced.}
  \label{fig:r_0-vs-R_m}
\end{figure}

While the geometric analysis above and in~\cite{FB-MG-RN:21} are a good start, the complications stemming from the radius~$R_m$ and the macrostates themselves being a truly emergent property makes it difficult to use in identifying rigid bounds of operation. For instance, although the proof above shows that the regular polygon in Fig~\ref{fig:iscoceles_triangle}(a) will remain stable if undisturbed, we are unable to show that it is a globally asymptotically convergent set (i.e. no matter the initial conditions, given the specific parameters, the agents will always form a perfect circle). Through simulating general setups where they don't already begin in a circle, we found that although simulations with parameters in $\PP*$ often do form this perfect mill eventually, there were rare cases near the boundary of this critical parameter space where eventually they would fall apart into the unorganized phase as can be seen in Fig~\ref{fig:r_0-vs-R_m}. For now we can only acknowledge this region of uncertainty, the grey band in Fig~\ref{fig:r_0-vs-R_m}, and are unable to make any claims about the exact size.

We propose following a route similar to the history of fluid mechanics. The Moody diagram Fig.~\ref{fig:moody-diagram} shows empirical relations between the dimensionless Reynolds number and the friction factor for varying `roughness'. While we one day hope to have a closed-form equation, we can see that even without exact equations a better understanding of how~$R_m$ depends on the different parameters may be possible by similarly analyzing only certain regimes and providing piece-wise equations based on empirical data. Even the Reynolds number is not always able to predict correctly if a flow is actually laminar or turbulent, but clearly has been good enough to allow us to control not only which macrostate water is in, but also the properties of water in the different phases. The hope is that this can allow engineers to start actually using collective behaviors of swarms even without a full understanding of how the collective properties emerge. Here we've started with a few empirical relations we have found and hope it motivates others to follow a similar path to deploy emergent behaviors in different starting pockets of predictability, and working to expand from the known areas to the unknown areas of the phase space of this system. 

Thus, Proposition~\ref{prop:at_manifold} provides sufficient but not necessary conditions for creating a milling circle with the milling radius of $R_m$. Furthermore, these results are only providing an example of a set that will create a perfect mill of $\overline{c} = 0$ (although as shown in Table~\ref{tbl:macrostates1}, we define a mill to be anything with $\overline{c}<0.3$); we are yet to establish the conditions that do lead to milling in the general case.

%Additionally, we are \textbf{not} claiming that if the system does not satisfy the conditions identified in Proposition~\ref{prop:at_manifold}, then the system will not have a specific milling radius $R_m$ or that it won't be in the milling macrostate at all. 

\subsection{Applying Swarm Mechanics}
Equipped now with a way to visualize the entire parameter space of sufficient conditions that can give rise to the milling behavior, we leverage Proposition~\ref{prop:at_manifold} to deploy an emergent milling behavior with a milling radius of our choosing~$R^*_m > 0$.

%Although these results do not provide the full solution to Problem~\ref{pr:mech}, they do allow us to instead answer the alternative of this question. Using the conditions and the equation for the milling radius found in Proposition~\ref{prop:at_manifold}, we can now provide infinite choices of how to setup the parameters $P$ of a system with kinematics~\eqref{eq:real_func} running the binary controller~\eqref{eq:control} to guarantee that the swarm will produce the milling macrostate with any desired $R_m$.

In other words, given the target $R^*_m$, we can immediately visualize the space of parameters that can support this milling radius by computing the set of all parameters that produce the mill with radius~$R^*_m$,

\begin{align}\label{eq:DEPLOYHERE}
R^{*(-1)} \triangleq \{P \in \mathcal{P} | \phi = \frac{2\pi}{N}, \frac{v}{\omega} \leq R^*_m, \gamma = 2R^*_m\sin(\frac{\pi}{N}) \}
\end{align}

% $R^{*(-1)} \triangleq \{P \in \mathcal{P} | \phi = \frac{2\pi}{N}\} \cap \{P \in \mathcal{P} | \gamma = 2R^*\sin(\frac{\pi}{N}) \} \cap
% \{P \in \mathcal{P} | \frac{v}{\omega} \leq R^* \}
% $

\begin{figure}[t]
\centering
  \includegraphics[width=.45\linewidth]{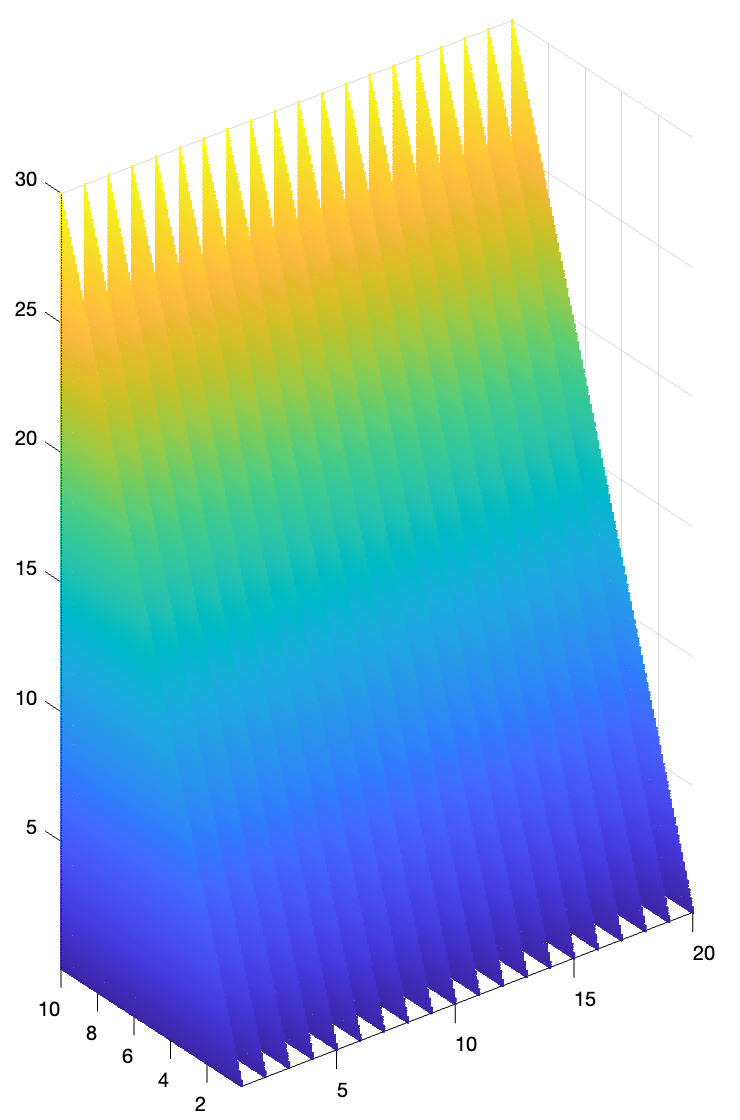}
  \put(-138,70){$v(\frac{m}{s})$}
  \put(-120,0){$\omega(\frac{rad}{s})$}
  \put(-30,0){$N$}
  \caption{A 3D point cloud visualizing the set that guarantees the system to be in the $M$ macrostate with the specified $R_m^*$, as long as $\phi = \frac{2\pi}{N}$ and $\gamma = 2R^*_m\sin(\frac{\pi}{N})$.}
  \label{fig:R_star_point_cloud}
\end{figure}

This set can be visualized in Fig~\ref{fig:R_star_point_cloud} where the desired $R^*_m = 3$. If  $\phi = \frac{2\pi}{N}$ and $\gamma = 2R^*_m\sin(\frac{\pi}{N})$, then any point selected from this point cloud is able to maintain a milling circle of radius $R^*_m$.

\section{Robot Validation}\label{se:validation}

Our ultimate goal with the novel framework is to use both simulations and new swarm chemistry/mechanics ideas to deploy real robot swarms with predictable emergent properties. However, any robots used will undoubtedly be a discrete system that can only sense and act at a certain rate, and thus the time sampling may play a factor that we have ignored given our use of a continuous-time model and therefore the results of the real robot experiment may be slightly different. However, by following the methods of our companion work on establishing a low-fidelity simulator using our proposed real2sim2real process for swarms~\cite{RV-KZ-SL-MP-CN:23}, we can shrink this gap enough such that we should be able to observe the behaviors found in simulation on the real robots used called Flockbots~\cite{SL-KA-MB-DF-KS-BH-CV-AB-RS-BD:14}.

\begin{figure}[h]
\centering
  \includegraphics[width=.7\linewidth]{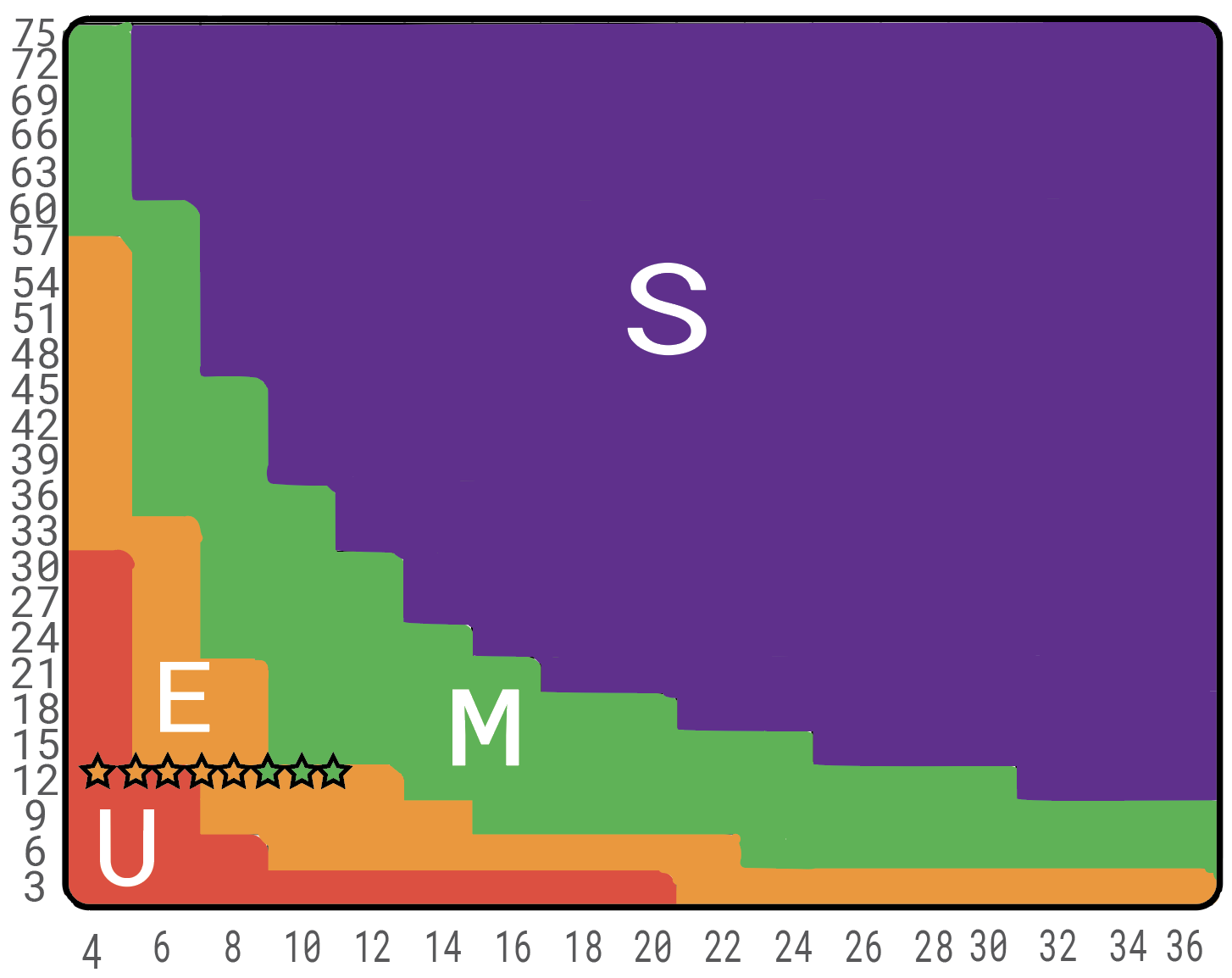}
  \put(-187,70){$\phi$}
  \put(-80,-10){$N$}
  \caption{A 2D phase diagram constructed in simulation overlaid with markings (stars) showing the impact of $N$ and $\phi$ (in degrees) on the behavior of simulated agents and real Flockbots.}
  \label{fig:Real_phase_diagram}
\end{figure}

Naturally it is easier to tune various parameters in simulation than with real hardware. For instance it is not easy to simply tune the FOV angle~$\alpha$ of the IR sensors built into the Flockbots and are limited to only being able to use up to~$N=11$ of the robots. Fig.~\ref{fig:Real_phase_diagram} shows our simulated phase diagram with a sequence of real robot experiments shown at~$\phi = 12^\circ$ and~$N = 4-11$ agents. This phase diagram and swarm mechanics analysis above allows us to make a hypothesis that ~$N = 9$ robots are needed before we see well-shaped milling circles (which is verified with the real robot experiments), and the size of the circle will keep increasing as we add robots until~$N=31$ robots when swarm chemistry tells us the circle will be too large to sustain and break out into smaller components.

In the future, we hope to be able to run an experiment with $N > 30$ robots to verify if the system will produce the multiple sub-group $S$ behavior. Furthermore, following our proposition~\ref{prop:at_manifold} given the limited $\phi$, we predict that we would need 30 robots to be able to apply the calculated milling radius equation~\eqref{eq:R_m} to verify its accuracy as well.

\begin{figure}[h]
\centering
  \includegraphics[width=.4\linewidth]{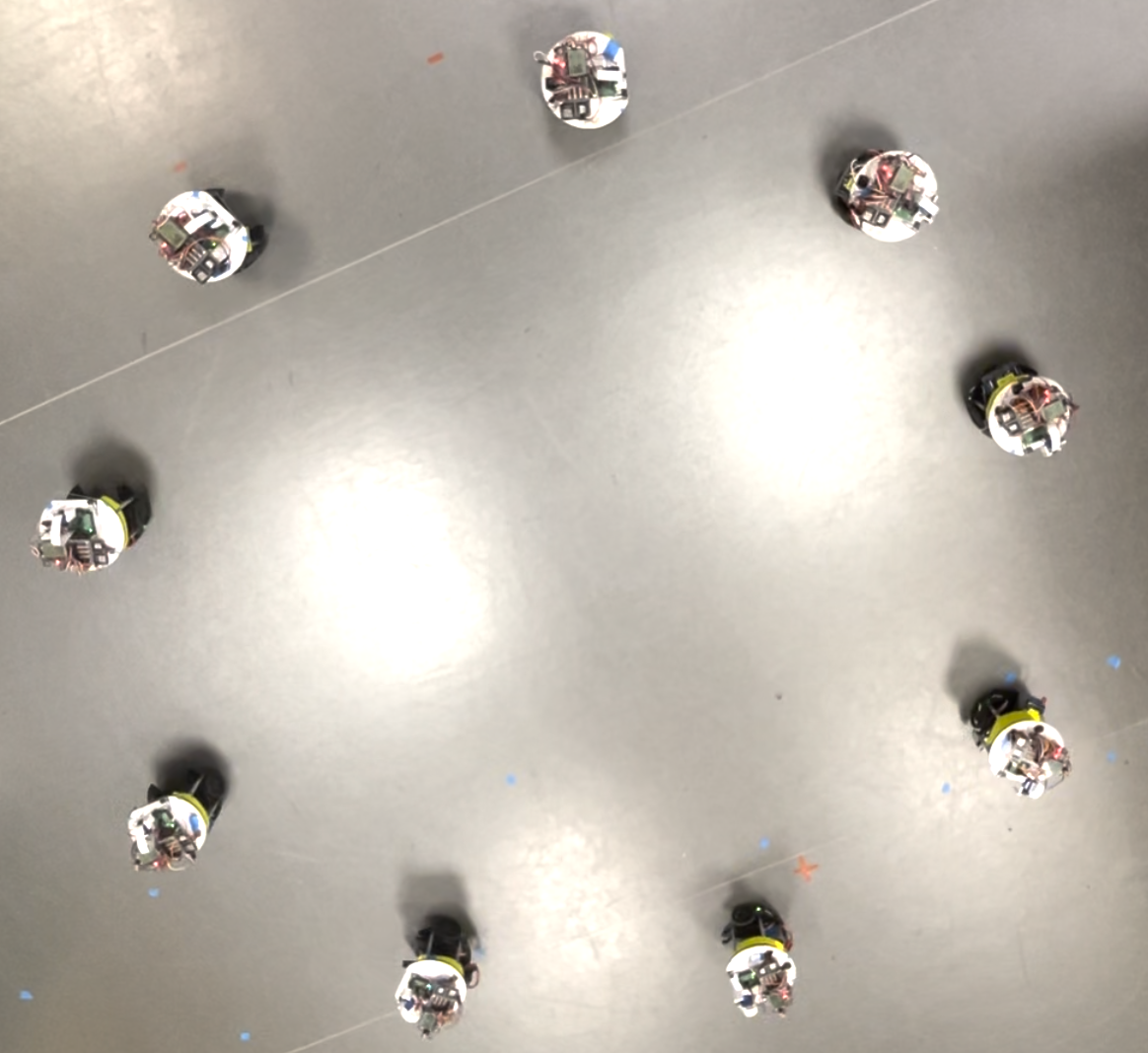}
  \caption{Nine Flockbots~\cite{SL-KA-MB-DF-KS-BH-CV-AB-RS-BD:14} in the milling macrostate.}
  \label{fig:Real_robots_milling}
\end{figure}

\begin{figure}[h]
\centering
  \includegraphics[width=.75\linewidth]{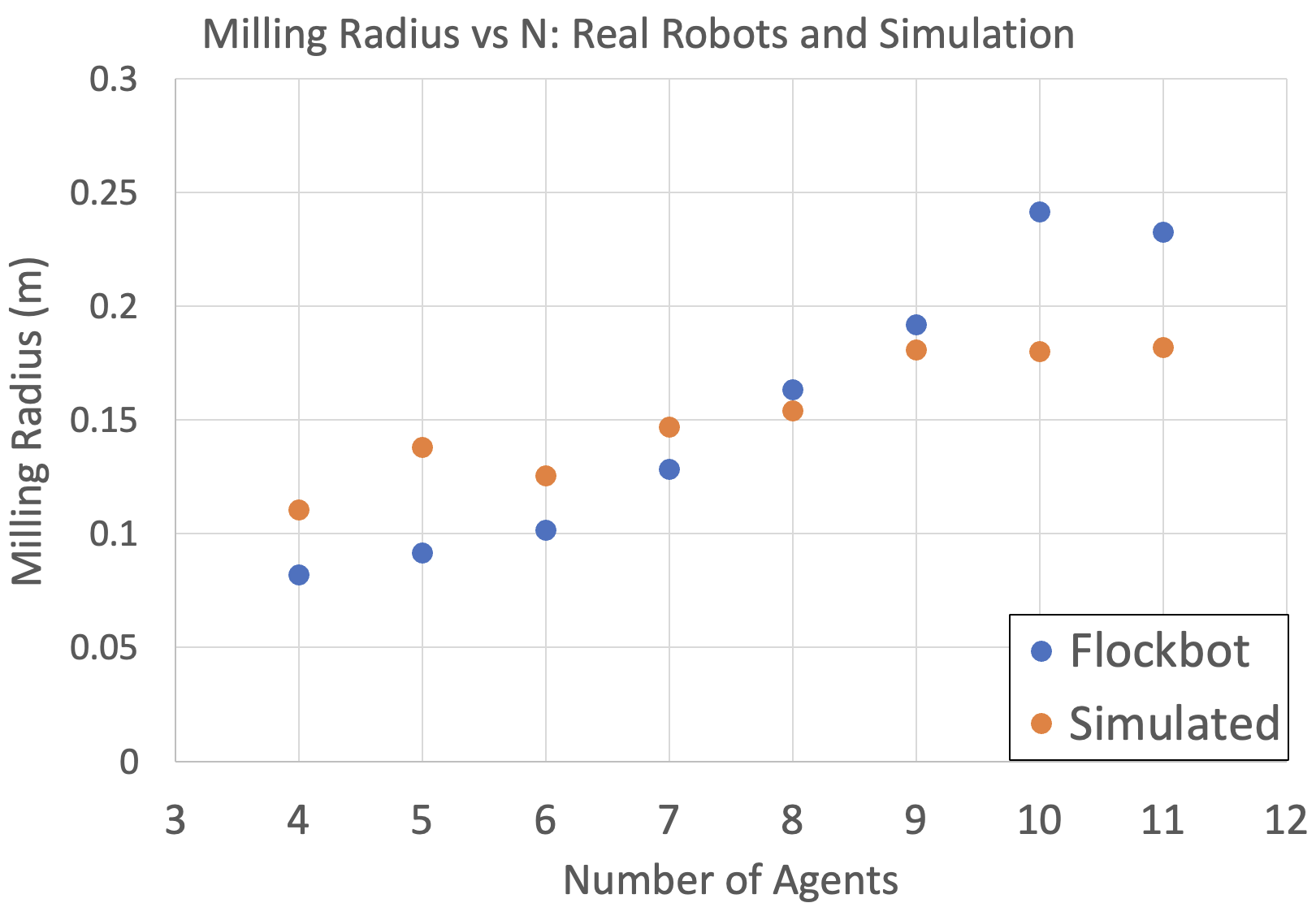}
  \caption{Radius of milling circle formed with various number of agents in our simulator and real robot experiments conducted on the Flockbots system}
  \label{fig:Real_vs_sim_vs_theoretical}
\end{figure}

\section{Conclusions}\label{se:conclusions}

This paper proposes a novel approach for analyzing and modeling swarms by turning to chemistry and fluid mechanics for inspiration. The primary contributions in this work are the novel framework and the formulation of new open questions that we believe deserve investigation from the swarms community at large. Our preliminary solutions are naturally incomplete and the goal instead is to draw more attention to these problems and new way of thinking about swarms. Future work will be devoted to seeking more complete solutions to Problems~\ref{pr:chem} and~\ref{pr:mech}. Our results are validated through both simulations and real experiments on ground robots.

%It should be noted that while we applied our novel framework to a simple swarm achieving a specific (already known) behavior, the methods and problems posed can easily be generalized to other swarming phenomena as long as proper swarm-level metrics are defined. 

\section*{Acknowledgments}
This work was supported in part by the Department of the Navy, Office of Naval Research (ONR), under federal grant N00014-22-1-2207.

\bibliographystyle{ieeetr}
\bibliography{ricardo}

\appendix

\textbf{Proof of Proposition~\ref{prop:at_manifold}}

Letting~$r_\text{max} = \max_{i \in N} ||\mathbf{p}_i-\mu||$ and~$r_\text{min} = \min_{i \in N}||\mathbf{p}_i-\mu||$, we can write 
\begin{align}\label{eq:circly}
\overline{c} = \frac{\max_{i \in N} ||\mathbf{p}_i-\mu|| - \min_{i \in N}||\mathbf{p}_i-\mu||}{\min_{i \in N}||\mathbf{p}_i-\mu||} = \frac{r_\text{max}}{r_\text{min}} - 1
\end{align}

\begin{align*}
     \dot{\overline{c}} = \frac{\partial \overline{c}}{\partial Z} \cdot \dot{Z} = \frac{r_\text{min} \dot{r}_\text{max} - r_\text{max} \dot{r}_\text{min}}{r_\text{min}^2} 
     %\leq \frac{1}{m^2} (m |\dot{M}| + M |\dot{m}|)
\end{align*}

The proof requires that the set of states such that~$r_\text{min} = r_\text{max} = R_m$ (and thus~$\overline{c}(t) = 0)$ is a positively invariant set. To prove invariance, we show that if~$\overline{c}(t) = 0$ for all~$t \geq 0$, then it must be that~$r_\text{min} = r_\text{max} = R_m$. From~\eqref{eq:circly}, it is clear that the only way for~$\overline{c}(t) = 0$ is if and only if~$r_\text{max}(t) = r_\text{min}(t)$

\begin{figure}[h]
    \centering
    \subfigure[]{\includegraphics[width=.55\linewidth]{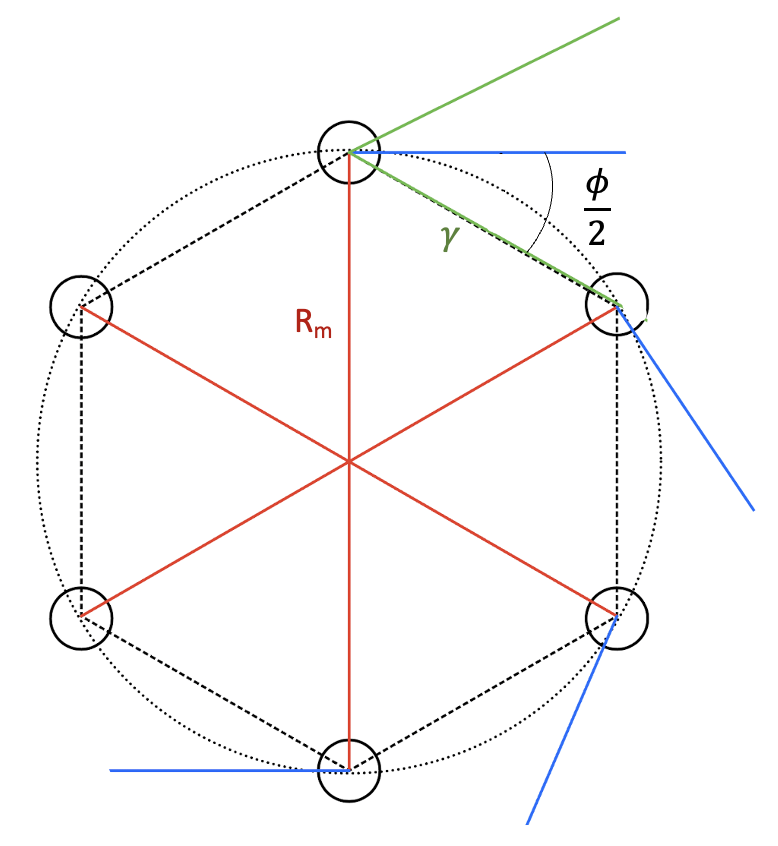}}\\
    \subfigure[]{\includegraphics[width=.35\linewidth]{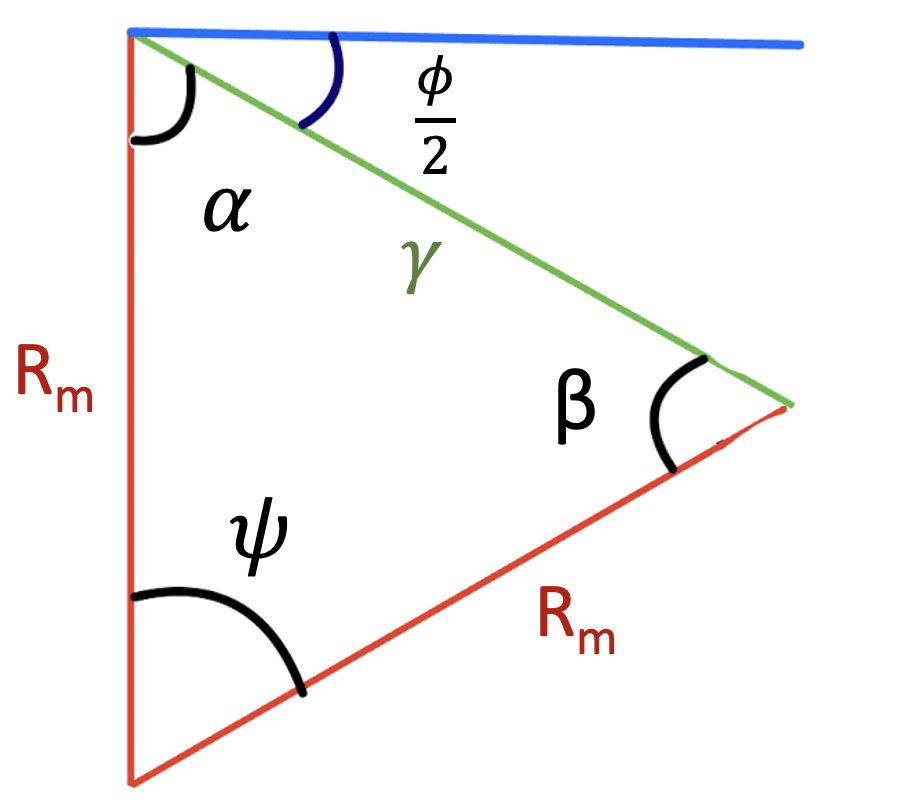}}
    \subfigure[]{\includegraphics[width=.35\linewidth]
    {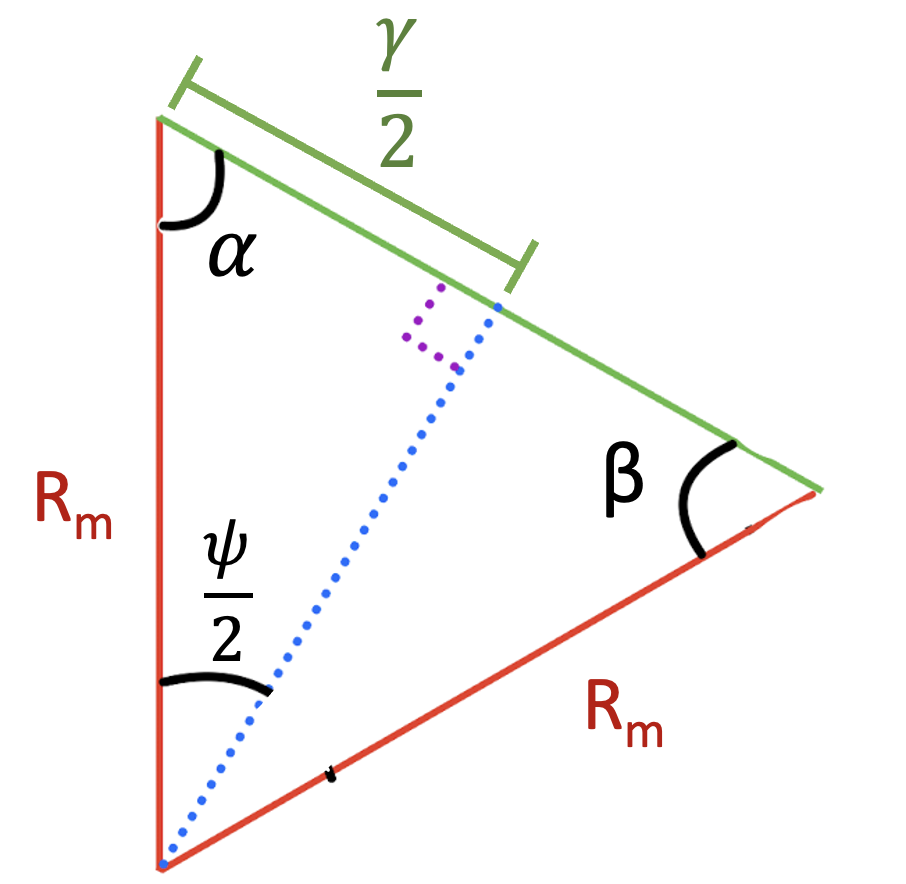}}
    \caption{(a) Regular polygon inscribed the milling circle of a 6-agent system. The blue lines represent the heading of the agents, while the green lines show the region of detection for the top agent.(b) Zoomed in triangle formed between top two agents (c) isosceles triangle broken down into more detail.}
    \label{fig:iscoceles_triangle}
\end{figure}

Let us now further analyze the set of states where~$\overline{c}(t) = 0$ geometrically. In the case that~$\dot{r}_\text{min}(t) = \dot{r}_\text{min}(t) = 0$, all the agents are equidistant from the average position of the agents (center of the mill) and are positioned at the vertices of a regular $N$-sided polygon (Fig \ref{fig:iscoceles_triangle}(a)). For~$\dot{r}_\text{max}(t) = \dot{r}_\text{min}(t) = 0$, the heading of all the agents should, on average, be tangential to the milling circle.  

% \begin{figure}[h]
% \centering
%   \includegraphics[width=.5\linewidth]{images/inscribed_polgyon_proof1.png}
%   \caption{Regular polygon inscribed the milling circle of a 6-agent system. The blue lines represent the heading of the agents, while the green lines show the region of detection for the top agent.}
%   \label{fig:inscribed_polygon_proof1}
% \end{figure}

Looking at this regular polygon, the triangle formed by the top agent and the one to its right can be seen in Fig~\ref{fig:iscoceles_triangle}(b), where it is clear to see that the triangle has two equal sides of length $R_m$ (i.e. isosceles triangle) therefore we know $\alpha = \beta$. The central angle of a regular polygon is $\frac{2\pi}{N} = \psi$ and using the sum of interior angles for a triangle we calculate $\alpha = \frac{\pi}{2} - \frac{\pi}{N}$. 

% \begin{figure}[h]
% \centering
%   \includegraphics[width=.25\linewidth]{images/iscoceles_triangle1.png}
%   \caption{Zoomed in triangle formed between top two agents.}
%   \label{fig:iscoceles_triangle1}
% \end{figure}

% \begin{figure}[h]
% \centering
%   \includegraphics[width=.25\linewidth]{images/iscoceles_triangle_more.png}
%   \caption{Iscoceles triangle broken down into more detail.}
%   \label{fig:iscoceles_triangle_more}
% \end{figure}

We can break this isosceles triangle down into two right triangle as shown in Fig:\ref{fig:iscoceles_triangle}(c). Then we can simply use trigonometry to solve for $R_m$:

\begin{align*}
\sin(\frac{\psi}{2}) &= \frac{\frac{\gamma}{2}}{R_m}\\
R_m &= \frac{\gamma}{2\sin(\frac{\psi}{2})} \\
R_m &= \frac{\gamma}{2\sin(\frac{\pi}{N})}
\end{align*}

Since the agents are assumed to be positioned on the vertices of a regular polygon and $\alpha$ makes a right angle with the half-FOV then we can determine that for this polygon formation to be stable then the FOV $\phi = \frac{2\pi}{N}$. More specifically, the stable mill of radius $R_m$ is only guaranteed for the critical value~$\phi = \frac{2\pi}{N}$.

Looking at Fig~\ref{fig:iscoceles_triangle}(a), we see that the individual turning radii ($\frac{v}{\omega}$) must not be larger than the $\frac{\gamma}{2\sin(\frac{\pi}{N})}$, the radius of the circumscribing circle. Otherwise the agents may travel along their individual turning radius, never converging to a stable regular polygon. 

\begin{figure}[h]
\centering
  \includegraphics[width=.35\linewidth]{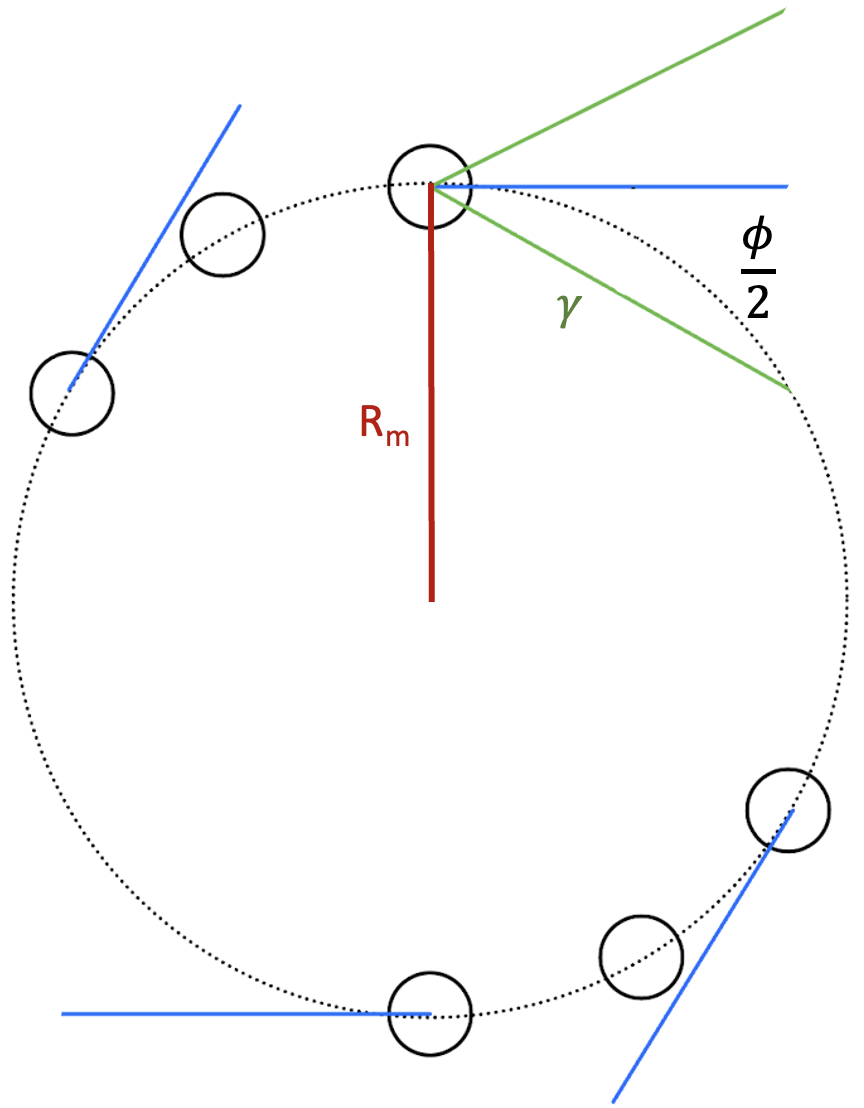}
  \put(-45,100){\tiny A }
  \put(-45,0){\tiny B }
  \caption{Agent configuration where $m = M = R_m$ is not a positively invariant set.}
  \label{fig:counter_example}
\end{figure}

It may also be argued that an alternative configuration (Fig~\ref{fig:counter_example}) may also satisfy~$r_\text{min}(t) = r_\text{max}(t)=R_m$ (and thus~$\overline{c}(t) = 0)$. In this case, Agents A and B may move along the circular paths set by their turning rate ($\frac{v}{\omega}$), which according to our defined $\PP^*$, can be set equal to $R_m$. However, the other agents in this configuration will start turning in the counter-clockwise circular path until they no longer detect the agent ahead. This will ultimately lead to a changing $r_\text{max}$, therefore, for a system with the specified controller~\eqref{eq:control}, this configuration is not a stable one.

Therefore, the only formation where~$r_\text{min} = r_\text{max} = R_m$ (and thus~$\overline{c}(t) = 0)$ is a positively invariant set is Fig~\ref{fig:iscoceles_triangle}(a) where the stable milling radius can be found with equation~\eqref{eq:R_m}.

\end{document}